\documentclass[runningheads]{llncs}

\newif\ifarxiv
\arxivtrue

\usepackage{eccv}

\usepackage{eccvabbrv}

\usepackage{graphicx}
\usepackage{booktabs}
\usepackage{xcolor}      
\usepackage{multirow}    
\usepackage{colortbl}    
\usepackage{array} 
\newcolumntype{C}[1]{>{\centering\arraybackslash}p{#1}} 

\usepackage[accsupp]{axessibility}  

\ifarxiv
  \usepackage[colorlinks,citecolor=eccvblue]{hyperref}
\else
  \usepackage{hyperref}
\fi

\usepackage{orcidlink}

\usepackage{amsmath,amsfonts,bm}

\def\eqref#1{equation~\ref{#1}}

\def\1{\bm{1}}

\def\vtheta{{\bm{\theta}}}

\def\vc{{\bm{c}}}

\def\vp{{\bm{p}}}

\def\vu{{\bm{u}}}
\def\vv{{\bm{v}}}

\def\vx{{\bm{x}}}

\def\vz{{\bm{z}}}

\def\mI{{\bm{I}}}

\def\mM{{\bm{M}}}

\DeclareMathAlphabet{\mathsfit}{\encodingdefault}{\sfdefault}{m}{sl}
\SetMathAlphabet{\mathsfit}{bold}{\encodingdefault}{\sfdefault}{bx}{n}

\newcommand{\model}{DriveWeaver}

\begin{document}

\title{\model{}: Point-Conditioned Video Inpainting for Controllable Vehicle Insertion in Autonomous Driving Simulation} 

\titlerunning{\model{}}


\author{%
    Junzhe Jiang\inst{1}
    \quad 
    Zipei Ma\inst{1,2}
    \quad 
    Zijie Pan\inst{1}
    \quad
    Li Zhang\inst{1,2}\thanks{Corresponding author (\url{lizhangfd@fudan.edu.cn})}
}

\ifarxiv
  \institute{School of Data Science, Fudan University \and
  Shanghai Innovation Institute \\
  \href{https://github.com/LogosRoboticsGroup/DriveWeaver}{\texttt{github.com/LogosRoboticsGroup/DriveWeaver}}\\
  }
\else
  \institute{School of Data Science, Fudan University \and
  Shanghai Innovation Institute \\
  \href{https://github.com/LogosRoboticsGroup/DriveWeaver}{\texttt{github.com/LogosRoboticsGroup/DriveWeaver}}
  }
\fi

\authorrunning{J.~Jiang et al.}

\maketitle


\begin{abstract}

A pivotal step in autonomous driving simulation involves inserting foreground vehicles with predefined trajectories into simulated scenes. This process enhances scene diversity and facilitates the creation of various corner cases for testing and improving autonomous driving models. However, existing methods often rely on pre-reconstructed 3D assets, which frequently lead to lighting inconsistencies between the inserted foreground and the background. Moreover, the reliance on limited, manually-curated 3D assets hinders large-scale deployment. To address these challenges, we propose \textbf{\model{}}, a novel framework for controllable vehicle insertion in autonomous driving simulation. Specifically, for a masked target insertion area, \model{} performs video inpainting conditioned on vehicle point clouds to generate high-quality, temporally consistent vehicles. This video-inpainting-based approach ensures seamless blending between the foreground and background, while the readily available point cloud conditions enable superior generalization. To support long-term generation, we further design a global-to-local hierarchical inpainting strategy, ensuring the consistent identity and appearance of the inserted vehicles. 
Meanwhile, we extract explicit 3D Gaussian representations of the inserted vehicles through an urban reconstruction pipeline to enable real-time rendering for autonomous driving simulation.
Extensive experiments across diverse datasets demonstrate that our method outperforms existing baselines in visual realism and geometric consistency, providing a robust tool for scalable autonomous driving scene augmentation.

\keywords{Video inpainting \and Autonomous driving simulation}

\end{abstract}   
\section{Introduction}
\label{sec:intro}

Autonomous driving simulators have garnered significant attention due to their ability to evaluate driving algorithms in a safe and cost-effective manner~\cite{dosovitskiy2017carla, yang2023unisim, chen2025snerf, ljungbergh2024neuroncap, realengine, zhou2024hugsim}. Traditional simulators~\cite{dosovitskiy2017carla, airsim2017fsr} employ manually-crafted CAD models to render autonomous driving scenarios. While offering high controllability, these simulators suffer from a pronounced domain gap compared to the real world. Recently, leveraging readily available real-world urban imagery, scene reconstruction techniques have provided a reliable solution for high-fidelity autonomous driving simulation. With advancements in neural rendering~\cite{xie2023snerf, yang2023emernerf, chen2023periodic, yan2024street, chen2024omnire, Ma2025BezierGS} which further augmented by generative priors to refine novel view synthesis~\cite{yan2024streetcrafter, zhao2024drivedreamer4d, yang2024drivex}, contemporary methods have achieved high-quality, large-scale novel view synthesis of recorded scenes. Despite 
achieving photo-realistic rendering, these methods are primarily restricted to faithfully reproducing captured environments. The lack of the capability for flexible scene editing poses a significant barrier to evaluating driving algorithms in rare, long-tail corner cases.

\begin{figure}[t]
    \centering
    \includegraphics[width=\linewidth]{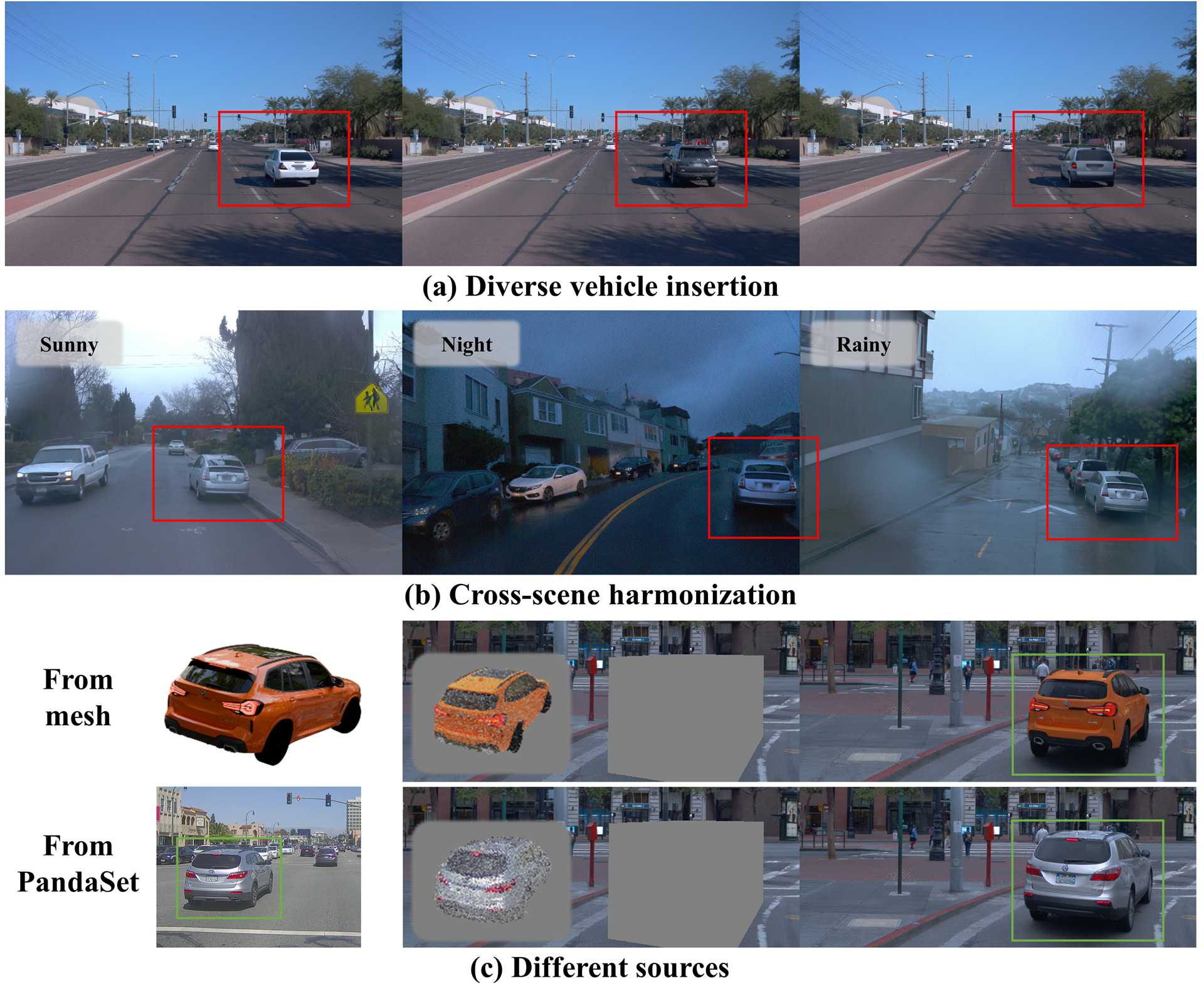}
    \caption{\textbf{Controllable high-fidelity vehicle insertion via \model{}.} Our point-conditioned inpainting framework provides flexible scene editing for autonomous driving simulation. It supports (a) diverse vehicle insertion options, (b) seamless insertion of vehicles with lighting in complex weather and illumination conditions and (c) flexible point-conditioned insertion using point clouds from diverse sources.
    }
    \label{fig:teaser}
\end{figure}

To achieve high-fidelity foreground insertion for constructing safety-critical corner cases, several recent studies~\cite{xie2023snerf, zhou2024hugsim, zhou2025nexus} leverage large-scale autonomous driving datasets~\cite{du20243drealcar, waymo, nuplan} to pre-reconstruct 3D vehicle assets. These assets are then composed and rendered into background scenes according to predefined trajectories. To mitigate the visual inconsistencies between the inserted objects and the environment, various post-processing techniques have been employed, such as selecting vehicles with matching illumination and synthesizing shadows. Despite these efforts, a persistent stylistic gap remains between the foreground and background. Furthermore, these pipelines heavily rely on labor-intensive processes, including manual pose calibration for unannotated datasets~\cite{du20243drealcar} and the heuristic selection of lighting-compatible assets. Such dependencies on manual intervention significantly limit the scalability and generalizability  across diverse real-world scenarios.
Alternative approaches~\cite{dipir, chen2025snerf, lu2024urbancad, realengine} employ manually-modeled 3D mesh assets as foreground objects and utilize diffusion models~\cite{sd} as evaluators to provide guidance signals for physics-based inverse rendering. Through differentiable rendering, these methods enable the end-to-end optimization of lighting parameters. While achieving superior realism in vehicle insertion, these frameworks require per-scene optimization via diffusion models, which is computationally expensive. Furthermore, the acquisition of high-quality 3D meshes remains costly, making these methods difficult to adapt efficiently to diverse scenarios or to satisfy the requirements of large-scale deployment.

In this paper, we present \textbf{\model{}}, a novel point-conditioned video inpainting diffusion model. As shown in~\cref{fig:teaser}, our framework achieves precise control over the synthesis of high-quality, geometrically consistent foreground vehicles using cost-effective point cloud data. Specifically, we leverage point cloud renderings as pixel-level conditions to provide rigorous geometric grounding for vehicle poses. By integrating these renderings as conditional inputs, \model{} exploits the powerful generative priors of video diffusion models to fix missing structural details and ensure seamless environmental harmony. This pixel-level conditioning paradigm enables cross-dataset generalization during inference, allowing point clouds from diverse sources to guide the generation process. Consequently, our approach offers a scalable and versatile solution for foreground vehicle insertion in autonomous driving simulations. Furthermore, the superior geometric consistency of the synthesized videos enables the extraction of explicit 3D Gaussian representations for the inserted foreground vehicles via urban reconstruction pipelines. This capability facilitates real-time rendering, which is crucial for the large-scale training and evaluation of autonomous driving models.

Our main contributions are summarized as follows: 
\textbf{({\romannumeral 1})} 
We propose a novel video inpainting diffusion framework \model{} designed for highly scalable and generalizable foreground vehicle insertion in autonomous driving simulation.
Leveraging point cloud renderings as pixel-level conditions, \model{} generates foreground vehicles with strong geometric consistency while preserving environmental illumination and style coherence. 
\textbf{({\romannumeral 2})}
To support long-term generation, we develop a global-to-local hierarchical inpainting strategy that ensures the long-term temporal consistency of inserted foreground vehicles.
Furthermore, the generated results can be directly reconstructed into  3D Gaussian representations, supporting efficient rendering for downstream autonomous driving model training and evaluation.
\textbf{({\romannumeral 3})}
Extensive experiments conducted on two large-scale benchmark datasets (Waymo~\cite{waymo} and PandaSet~\cite{pandaset}) demonstrate that \model{} significantly outperforms existing state-of-the-art methods in terms of visual realism and geometric consistency.
\section{Related work}

\noindent{\bf Neural rendering for autonomous driving simulation.} 
Recent advancements in neural rendering have fundamentally transformed autonomous driving simulation by enabling high-fidelity, data-driven scene reconstruction directly from sensor logs. Early neural simulators~\cite{yang2023emernerf, nsg, turki2023suds, xie2023snerf, chen2025snerf} primarily relied on neural radiance fields (NeRF)~\cite{mildenhall2020nerf} to parameterize large-scale urban environments. To address the computational bottlenecks and representation limitations of NeRF, recent approaches~\cite{chen2023periodic, yan2024street, hugs, chen2024omnire, Ma2025BezierGS, yang2024storm} have increasingly adopted 3D Gaussian splatting (3DGS)~\cite{kerbl3Dgaussians} for highly efficient rendering.
Despite their remarkable reconstruction capabilities, these purely physical-based rendering methods often suffer from visual artifacts or blurring in unobserved regions when rendering extreme novel views, which frequently arise in autonomous driving simulation~\cite{hugs, zhou2025nexus, realengine, RAD}. 
To alleviate this, contemporary frameworks~\cite{wang2024freevs, yan2024streetcrafter, zhao2024drivedreamer4d, yang2024drivex} leverage the powerful generative priors of large-scale diffusion models to fix missing details and refine rendering quality. 
Although these neural rendering and generative frameworks achieve impressive realism when reconstructing recorded data, they lack the ability to edit scenes, which limits their applicability for evaluating driving algorithms in rare long-tail corner cases within autonomous driving simulators. This limitation motivates us to develop a point-conditioned video inpainting model that enables flexible editing of reconstructed scenes.

\noindent{\bf Urban scene editing.}
To test autonomous vehicles in safety-critical scenarios, simulators must support flexible scene editing, particularly foreground object insertion. Recent works~\cite{zhou2024hugsim, zhou2025nexus} curate large libraries of pre-reconstructed 3D vehicle assets and compose them with background scenes along predefined trajectories. To reduce lighting and style gaps, these pipelines rely on asset retrieval with matched illumination and shadows synthesis, which limits scalability and cross-scene generalization.
Other approaches~\cite{dipir, chen2025snerf, lu2024urbancad, realengine} adopt physics-based neural rendering~\cite{wang2023fegr} pipelines that explicitly optimize object-scene interactions using high-quality 3D meshes and inverse rendering. While capable of producing realistic results, these methods are computationally expensive and depend on costly CAD meshes, hindering large-scale deployment.
Concurrent works explore alternative editing paradigms that face similar scalability and temporal constraints. R3D2~\cite{r3d2} and G$^2$Editor~\cite{g2editor} compose foreground objects from pre-reconstructed 3D assets, which incur substantially higher acquisition costs than our point-based conditions. DriveEditor~\cite{driveeditor} and GenMM~\cite{genmm} instead rely on a single reference images for object insertion, which introduces self-occlusion artifacts and restricts insertion viewpoints to those matching the reference. Furthermore, these methods lack the temporal scale necessary for closed-loop simulation: R3D2 is image-only, while G$^2$Editor, DriveEditor, and GenMM are limited to approximately 10 frames---far below the 193 frames our approach generates. 
In contrast, \model{} replaces costly 3D assets and restrictive reference images with accessible point cloud renderings, enabling viewpoint-flexible, high-fidelity insertion in a single forward pass. Beyond data efficiency, our hierarchical global-to-local generation strategy eliminates long-term temporal drift, and 3DGS distillation bypasses per-frame inference bottlenecks, achieving real-time interactive rendering suitable for closed-loop autonomous driving simulation.

\noindent{\bf Video generation and inpainting.}
Video generation and inpainting techniques have advanced significantly, enabling diverse editing capabilities such as dynamic object insertion and missing region recovery~\cite{wang2019video, hu2020proposal, zou2021progressive, Xu_2019_CVPR, guo2023animatediff, propainter}. Recently, architectural improvements in Diffusion transformers (DiT)~\cite{Peebles2022DiT} and the development of flow matching~\cite{flow-matching, esser2024sd3} have led to a new generation of foundation video models~\cite{opensora, yang2024cogvideox, kong2024hunyuanvideo, wan, wanvace}. These models demonstrate unprecedented visual fidelity, scalability, and temporal consistency in video synthesis, benefiting various downstream tasks~\cite{lionlora, qiu2025cinescale, gao2025lora, zheng2026versecrafter}.
Building upon the strong generative capability of Wan2.1 VACE-14B~\cite{wanvace}, we formulate foreground vehicle insertion as a controlled video inpainting task. Specifically, we inject sparse point cloud renderings as explicit geometric conditions into the conditional branch of Wan2.1. This point-conditioned inpainting framework allows us to leverage the reliable texture and illumination priors of large-scale generative models, enabling the synthesis of high-quality foreground vehicles that are both geometrically consistent and harmoniously integrated with the autonomous driving scene.
\section{Method}
\label{sec:method}
\begin{figure}[t]
    \centering
    \includegraphics[width=\linewidth]{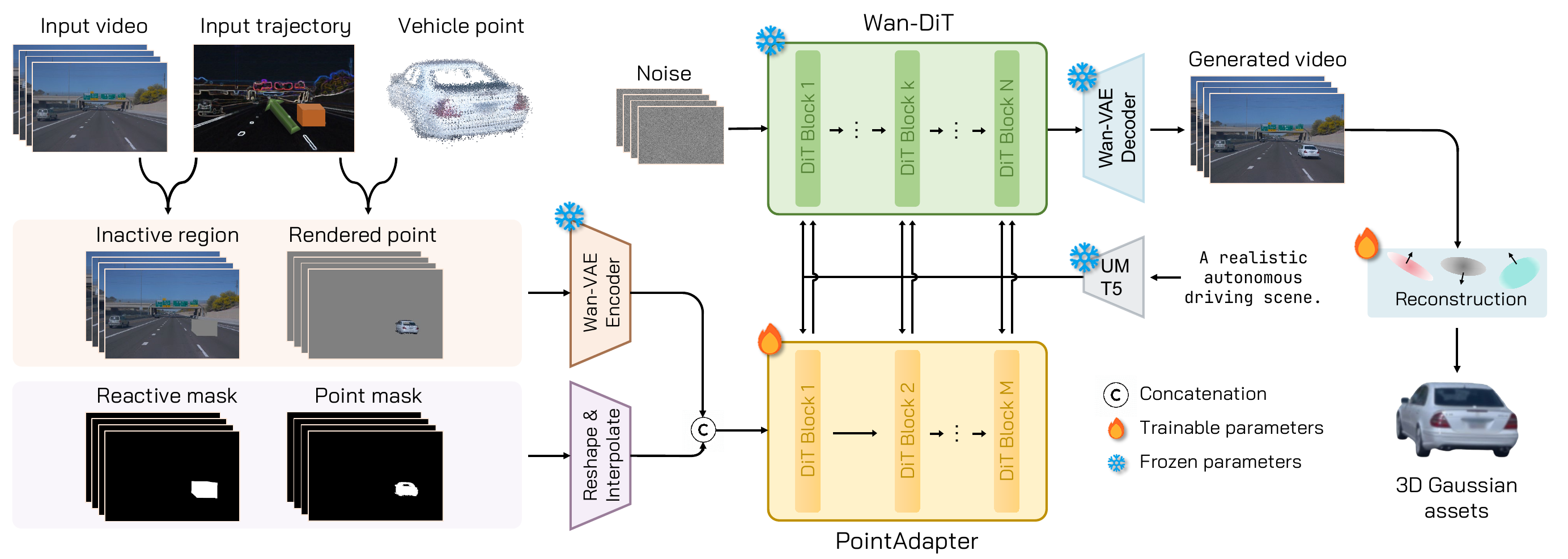}
    \caption{\textbf{Pipeline of \model{}.} Given an input driving video, a vehicle point cloud, and a target trajectory, \model{} generates realistic scenarios with the vehicle seamlessly inserted. The pipeline leverages a frozen video diffusion backbone (Wan-DiT) and introduces a lightweight, trainable PointAdapter. Specifically, the processed RGB conditions and masks are channel-wise concatenated and fed into the PointAdapter, which injects geometric and spatial guidance into the main branch via residual connections, ensuring efficient training and high-quality generation. The inserted vehicles can be further distilled into 3D representations using an urban reconstruction pipeline, enabling real-time rendering in autonomous driving simulation.
    }
    \label{fig:pipeline}
\end{figure}

Given a recorded autonomous driving scene with camera images, vehicle point clouds, and predefined vehicle trajectories, our goal is to synthesize high-fidelity vehicles that are seamlessly harmonized with the background and strictly aligned with the provided point clouds and trajectory constraints. 
In~\cref{sec:preliminary}, we briefly introduce flow matching models. Then, we introduce our point-conditioned video inpainting model \model{} in~\cref{sec:pipeline} and long-term generation strategy in~\cref{sec:long-term}. \cref{sec:distillation} describes how the inserted vehicles generated by DriveWeaver are distilled into 3D Gaussian representations to enable real-time rendering. Finally, \cref{sec:details} details the training dataset construction and implementation specifics. An overview of our pipeline is provided in~\cref{fig:pipeline}.

\subsection{Preliminaries}
\label{sec:preliminary}

\noindent{\bf Flow matching.} \model{} is built upon Wan2.1 VACE-14B~\cite{wanvace}, which employs the flow matching framework~\cite{flow-matching, esser2024sd3} to establish a consistent denoising diffusion process for diverse video distributions. 
An intermediate latent $\vx_t$ is constructed by interpolating between a video latent $\vx_1$ and Gaussian noise $\vx_0 \sim \mathcal{N}(0, I)$, where the timestep $t \in [0, 1]$ is drawn from a logit-normal distribution. Adopting the rectified flows~\cite{esser2024sd3} formulation, the latent at time $t$ is expressed as $\vx_{t} = (1 - t) \vx_{0} + t \vx_{1}$. The corresponding target velocity is defined as:
\begin{equation}\label{v_t}
\vv_{t} = \frac{\mathrm{d} \vx_{t}}{\mathrm{d} t} = \vx_{1} - \vx_{0}.
\end{equation}
The network is optimized to estimate this velocity field. Specifically, the objective function is defined as the mean squared error (MSE) between the predicted velocity $\vu(\vx_t, \vc_{txt}, t; \vtheta)$ and the ground truth $\vv_t$:
\begin{equation}\label{rf_loss}
\mathcal{L} = \mathbb{E}_{\vx_0, \vx_1, \vc_{txt}, t}||\vu(\vx_t, \vc_{txt}, t; \vtheta)-\vv_t||^2,
\end{equation}
where $\vc_{txt}$ represents the text embeddings extracted from UMT5~\cite{chung2023unimax} and $\vtheta$ denotes the trainable parameters.

\subsection{\model{} architecture}
\label{sec:pipeline}

This section focuses on developing a diffusion model for high-fidelity vehicle insertion. Given an $N$-frame sequence of recorded driving scenes, the vehicle point clouds, and their predefined trajectories, the model renders corresponding point cloud condition maps. It then generates a sequence of $N$ video frames with the inserted vehicles, thereby enabling the construction of rare long-tail scenarios for autonomous driving simulation. We employ Wan2.1 VACE-14B~\cite{wanvace} as our frozen latent video flow matching diffusion backbone, which comprises a 3D VAE and a diffusion transformer (DiT) denoiser. To enable the backbone to be conditioned on our rendered point cloud maps, \model{} introduces a lightweight point cloud adapter. This adapter modulates the frozen backbone, allowing it to incorporate the geometric guidance from the rendered point cloud maps during the generation process.

\noindent{\bf Point cloud rendering.} We employ a point cloud rendering technique to generate geometric conditions. For a driving sequence consisting of $N$ frames, we first project the LiDAR points onto the calibrated images and colorize them by querying the corresponding pixel values. We then extract the point clouds for each vehicle instance using 3D bounding boxes, resulting in instance point clouds $\{\vp_{j}^{i}\}_{j=1}^N$ defined in the canonical coordinate system of each vehicle instance $i$ in frame $j$. To render the point condition at frame $j$, we aggregate LiDAR points within a temporal window of length $l$. These points are integrated into a unified point cloud $\vp^i$ represented in the canonical coordinate system. Subsequently, the point cloud is transformed based on the respective vehicle instance poses and camera poses, and the point clouds from all individual instances are aggregated to form the combined point cloud $\vp$.

Since direct projection of the combined point cloud $\vp$ onto the image plane often leaves significant gaps and occluded regions, we assign a fixed radius to each point and perform point rasterization following~\cite{yan2024streetcrafter}. As illustrated in~\cref{fig:pipeline}, this procedure yields a set of rendered point cloud maps $\{\mI_{j}^{pc}\}_{j=1}^N$ and their associated masks $\{\mM_{j}^{pc}\}_{j=1}^N$. Simultaneously, the 3D bounding boxes of the inserted vehicles are projected onto the image plane to differentiate between regions requiring inpainting and the original background areas. This projection generates inactive background RGB images $\{\mI_{j}^{in}\}_{j=1}^N$, and binary inpainting masks $\{\mM_{j}^{re}\}_{j=1}^N$ which explicitly define the reactive area for vehicle synthesis. By using point cloud rendering as a pixel-level condition for video inpainting, the model only needs to complete the geometry indicated by the point cloud while maintaining environmental coherence. This formulation enables cross-dataset conditioning on vehicle point clouds and improves large-scale generalization.

\noindent{\bf PointAdapter.} We respectively concatenate the rendered point cloud maps $\{\mI_{j}^{pc}\}_{j=1}^N$ and their associated masks $\{\mM_{j}^{pc}\}_{j=1}^N$, the background RGB of the inactive regions $\{\mI_{j}^{in}\}_{j=1}^N$, and the masks for the reactive regions $\{\mM_{j}^{re}\}_{j=1}^N$. The two RGB maps ($\{\mI_{j}^{pc}\}_{j=1}^N$, $\{\mI_{j}^{in}\}_{j=1}^N$) are encoded using the same 3D VAE~\cite{wan} employed for the video latents. And following the practices in~\cite{wan, wanvace}, the masks ($\{\mM_{j}^{pc}\}_{j=1}^N$, $\{\mM_{j}^{re}\}_{j=1}^N$) are reshaped and interpolated to match the latent resolution. These components are then concatenated along the channel dimension and aligned with the latent video tokens.

Taking the concatenated conditions above as input, PointAdapter is designed as a lightweight DiT branch, as shown in~\cref{fig:pipeline}. It maintains the same token dimension as the Wan-DiT blocks but consists of significantly fewer layers. A PointAdapter block $\mathcal{T}^{pc}$ is associated with every $k$-th Wan-DiT block $\mathcal{T}$, and the output of the PointAdapter block is linearly projected back to the backbone's feature space by projection function $\mathcal{P}$ and incorporated into the corresponding DiT block as a residual modulation: 
\begin{gather}
\vz^{pc}_{j+1} = \mathcal{T}^{pc}_j(\vz^{pc}_{j}, \vc_{txt}, t)\\
\vz_{i+1} = \mathcal{T}_i(\vz_{i}, \vc_{txt}, t) + \mathcal{P}_{i/k}(\vz^{pc}_{i/k}) \cdot \mathbf{1}[\, i \equiv 0 \ \  (\mathrm{mod}\ k) \,]
\end{gather}
where $i$ and $j$ denote the block indices of the Wan-DiT and the PointAdapter, respectively. $\vz$ and $\vz^{pc}$ represent the latent features of the backbone and the PointAdapter. $\vc_{txt}$ refers to the text embeddings encoded by umT5~\cite{chung2023unimax}, $t$ is the timestep, and $\mathbf{1}[\cdot]$ is an indicator function that determines the specific backbone blocks where the geometric information is injected. This adapter-based conditioning mechanism injects geometric information into the backbone using only a minimal number of additional parameters, ensuring that the geometric guidance effectively affects each backbone layer while maintaining high training efficiency.

\noindent{\bf Inference.} During the inference stage, \model{} benefits from the pixel-level conditioning of the rendered point cloud maps, which enables support for cross-dataset and non-LiDAR point cloud inputs. To further enhance the generation quality, we employ classifier-free guidance (CFG)~\cite{cfg}. Specifically, in addition to the conditional branch, we define an unconditional input by zeroing out the rendered point cloud maps $\{\mI_{j}^{pc}\}_{j=1}^N$ and their corresponding masks $\{\mM_{j}^{pc}\}_{j=1}^N$. This guidance strategy allows \model{} to synthesize more realistic and consistent foreground vehicles within fewer sampling steps, effectively improving both visual fidelity and inference efficiency.

\subsection{Long-term generation}
\label{sec:long-term}
\begin{figure}[t]
    \centering
    \includegraphics[width=\linewidth]{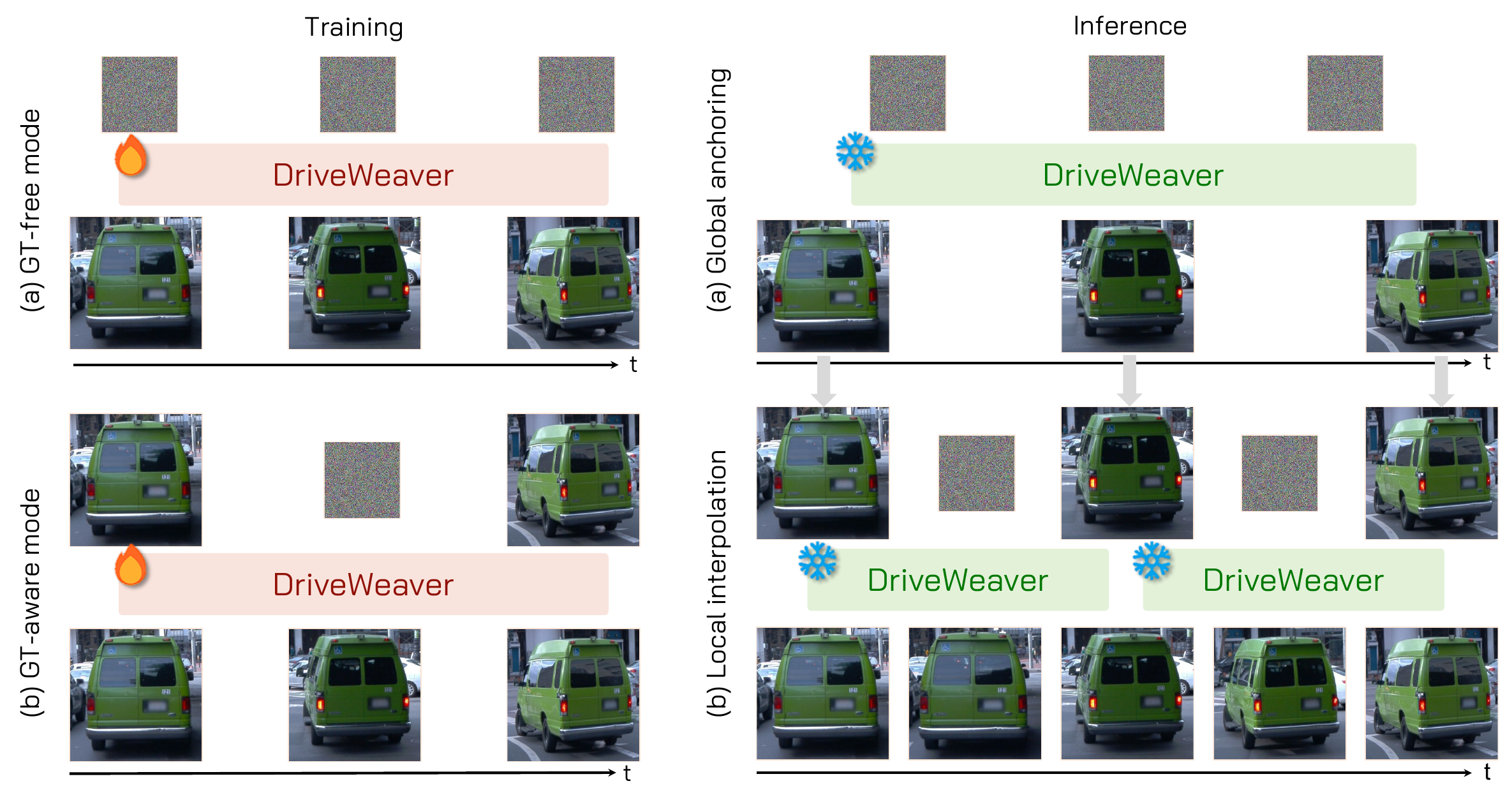}
    \caption{\textbf{Long-term generation.} To prevent error accumulation and semantic drift common in autoregressive methods, \model{} decouples the generation into two stages. (a) Global anchoring stage: The model processes a temporally subsampled sparse sequence in a single pass to establish a consistent vehicle identity and lighting across the entire trajectory. (b) Local interpolation stage: The synthesized sparse anchor frames are subsequently used as dense guidance to generate the local high-frame-rate segments, ensuring strict 3D and temporal consistency over extended durations. During training, \model{} is trained under GT-Free and GT-Aware modes to respectively accommodate the global anchoring and local interpolation stages.
    }
    \label{fig:long-term}
\end{figure}

In autonomous driving simulation, vehicle trajectories often span extended durations, necessitating the generation of long, temporally consistent videos. However, due to the high memory consumption of 3D-VAE and DiT architectures, modern video diffusion models are typically limited to generating short clips in a single pass (e.g., 13 to 81 frames). Existing approaches~\cite{gao2024vista, henschel2024streamingt2v, chen2023seine} often employ autoregressive or sliding window strategies, where the final frames of a previous clip serve as the condition for the next. While straightforward, this Markovian conditioning is prone to error accumulation. Over long durations, the generated foreground vehicles often suffer from significant semantic drift, structural degradation, and appearance inconsistency, which violates the strict 3D consistency required for driving simulation.
To overcome these limitations, we propose a global-to-local hierarchical inpainting strategy. Instead of sequential generation, we decouple the process into a global low-frame-rate anchoring stage and a local high-frame-rate interpolation stage, as shown in~\cref{fig:long-term}.

\noindent{\bf Global anchoring stage.} For a long input sequence, we first perform temporal subsampling on the background video and the corresponding point cloud conditions to create a sparse, low-frame-rate sequence. Given the reduced frame count, our model can process the entire duration in a single global forward pass. This step generates a coarse yet temporally consistent anchor video that establishes the global identity, geometry, and environmental lighting of the inserted vehicle across the full trajectory.

\noindent{\bf Local interpolation stage.} Subsequently, we partition the original high-frame-rate timeline into shorter local segments. For each segment, the corresponding anchor frames serve as dense guidance to replace the sparse point cloud conditions. The model then performs conditional generation to synthesize the local high-frame-rate video.

To support this two-stage inference paradigm, we train \model{} under two complementary modes. In the GT-Free mode, the model learns to generate frames solely from sparse point cloud conditions, preparing it for the global anchoring stage. In the GT-Aware mode, ground-truth frames are provided as supervision, enabling the model to perform precise local interpolation guided by the anchor frames.
By enforcing local generation to strictly follow the global anchors, our strategy effectively eliminates error accumulation and temporal drift. As a result, \model{} ensures that the appearance and lighting of the vehicle remain perfectly consistent from the first frame to the last.

\subsection{Distillation of inserted vehicles}
\label{sec:distillation}

While our point-conditioned video inpainting framework successfully synthesizes foreground vehicles with seamless environmental harmony and strict geometric consistency, the generative process is inherently bottlenecked by the slow, iterative nature of the denoising steps~\cite{ddpm, Peebles2022DiT}. For large-scale autonomous driving simulators that require high-frame-rate or real-time closed-loop evaluation~\cite{yang2023unisim, zhou2025nexus, zhou2024hugsim, realengine}, relying exclusively on a video diffusion model for real-time rendering is computationally prohibitive. To bridge the gap between high-fidelity offline generation and online simulation, it is imperative to distill the generated 2D video into an explicit 3D representation.
Because our hierarchical generation strategy and precise point-conditioning enforce temporal consistency, the synthesized video frames meet the geometric constraints required for 3D reconstruction.
Specifically, we leverage OmniRe~\cite{chen2024omnire}, a state-of-the-art 3D Gaussian~\cite{kerbl3Dgaussians} urban reconstruction framework, to process the synthesized high-quality video. During the reconstruction pipeline, we explicitly decouple the inserted vehicle instance from the background scene representations, which encapsulates the complex illumination, realistic shadows, and stylistic properties generated by the diffusion model. This distilled 3D asset can subsequently be integrated into 3D Gaussian rasterization pipelines. Consequently, during the simulation phase, we can completely bypass the computationally heavy diffusion model, enabling real-time rendering of the edited corner cases for autonomous driving simulation.

\subsection{Implementation details}
\label{sec:details}
We build our training dataset using the Waymo Open Dataset~\cite{waymo} and PandaSet~\cite{pandaset}, leveraging their 10Hz camera and the synchronized LiDAR. Following the instance point cloud rendering pipeline described in~\cref{sec:method}, we curated a total of approximately 10,000 training samples. We set the point rasterization radius to 0.005 for generating the rendered point cloud maps.
We construct \model{} based on the Wan2.1 VACE-14B~\cite{wanvace} model. The PointAdapter comprises 3B parameters and is integrated into the Wan-DiT backbone at layers $[0, 5, 10, \dots, 30, 35]$ (i.e., every 5th block). During training, the Wan backbone remains frozen, and only the PointAdapter modules are updated. 
To ensure that the model generates results solely conditioned on the rendered point cloud maps, we fix the text prompt as “A realistic autonomous driving scene”.
The model is trained on video sequences of 49 frames with a resolution of $480 \times 720$. We use the Adam optimizer with a learning rate of $1\times 10^{-5}$ and a total batch size of 8. A linear warmup of 100 steps is applied at the beginning of the training process, during which the learning rate gradually increases. All experiments are conducted on 8 NVIDIA H200 (141GB) GPUs, and the training process runs for 20,000 iterations (approximately 5 days) to complete. During training, we employ the classifier-free guidance (CFG)~\cite{cfg} strategy, where the point cloud conditions are randomly dropped with a probability of 0.15. At inference time, we use 30 denoising steps with a CFG scale of 2.5. Generating a 49-frame 480p video clip on a single H200 GPU takes approximately 2 minutes.

For long-term generation, we process sequences of 193 frames by sampling anchor frames at a stride of 4. This results in a 49-frame anchor sequence used to establish long-term consistency. The generated anchor video is then equally partitioned into 4 segments to facilitate local high-frame-rate synthesis. To distill the 3D representations of the inserted vehicles, we follow the configuration of OmniRe~\cite{chen2024omnire} and disable LiDAR depth supervision within the insertion regions, as the synthesized foreground vehicles lack precise depth ground truth.

Regarding computational overhead for distillation of inserted vehicles, training a standard 193-frame sequence requires approximately 1.5 hours on single NVIDIA RTX 4090 GPU. However, because OmniRe decouples the optimization of foreground instances from the background, we can freeze the pre-trained background 3DGS representation and exclusively optimize the inserted foreground vehicle. This targeted optimization converges in only 10 minutes per instance. Given that pre-reconstructed background 3DGS models serve as a standard prerequisite in practical autonomous driving simulation pipelines, this marginal offline distillation cost is readily justifiable.
\section{Experiments}
We evaluate both \model{} and its distilled 3D Gaussian representation (denoted as \model{} and \model{}-D, respectively) in terms of visual realism and geometric consistency. All evaluation metrics are computed after resizing the outputs to the training resolution of \model{}, i.e., $480 \times 720$.

\begin{figure}[t]
    \centering
    \includegraphics[width=\linewidth]{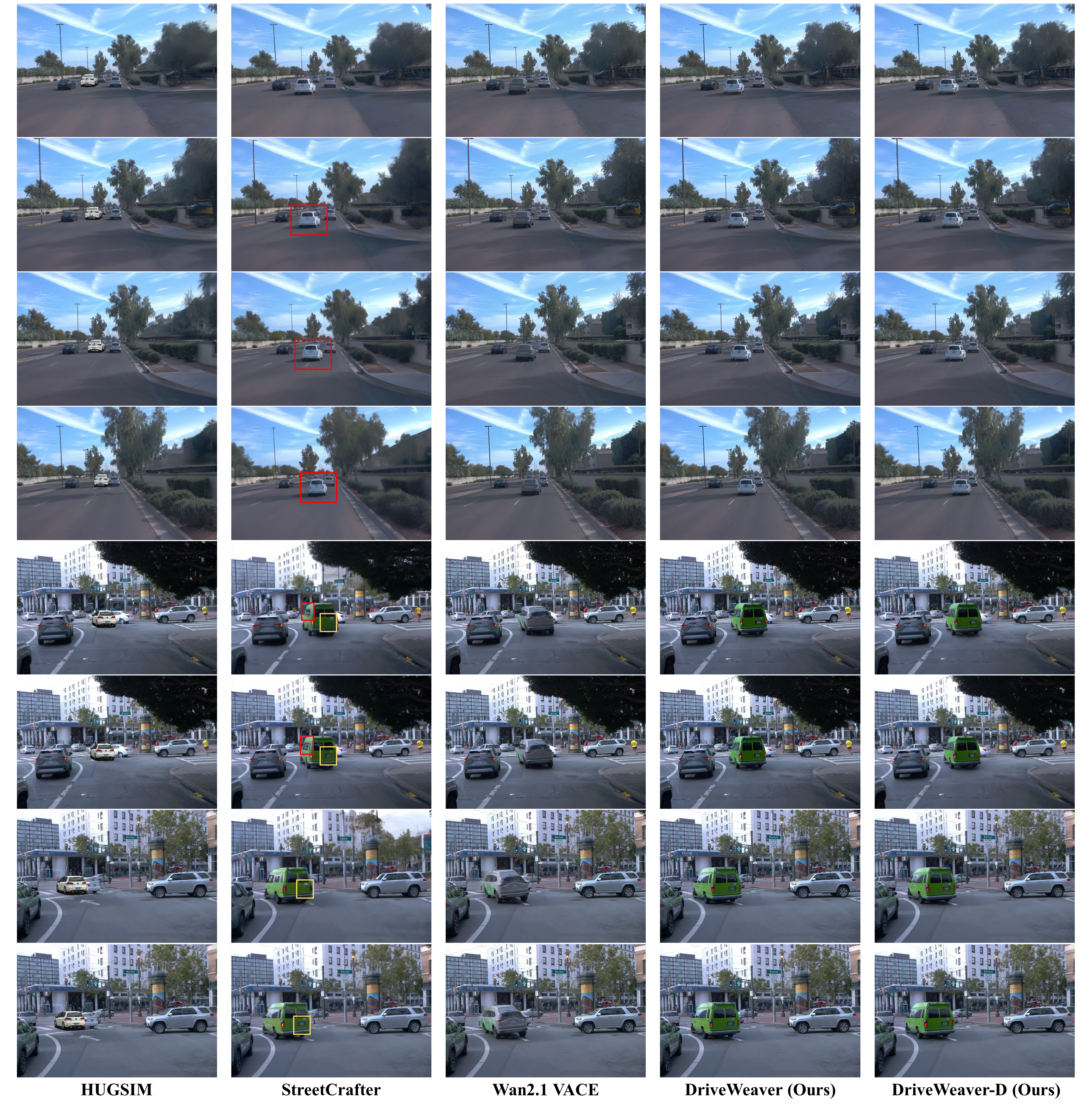}
    \caption{\textbf{Qualitative comparison on the Waymo~\cite{waymo} dataset.} 
    }
    \label{fig:waymo}
\end{figure}

\subsection{Experimental setup}
\noindent{\bf Baselines.} We compare our method with HUGSIM~\cite{zhou2024hugsim}, StreetCrafter~\cite{yan2024streetcrafter} and Wan2.1 VACE-14B~\cite{wanvace}.
HUGSIM~\cite{zhou2024hugsim} obtains 3D assets by reconstructing the 3DRealCar~\cite{du20243drealcar} dataset, combines them with the background for rendering, and renders shadows to enhance realism.
StreetCrafter~\cite{yan2024streetcrafter} takes LiDAR point cloud rendering as a condition to generate the entire scene.
And Wan2.1 VACE-14B~\cite{wanvace} is the backbone model we use, which supports the masked video to video (MV2V) generation mode.
All methods are based on their official implementations.

\noindent{\bf Evaluation metrics.} 
We conduct experiments on the Waymo Open Dataset~\cite{waymo} and PandaSet~\cite{pandaset}. Specifically, we select 10 sequences of approximately 200 frames each from the Waymo and 5 sequences of 80 frames each from PandaSet. All evaluation sequences are strictly held out and do not appear in the training set.
For visual realism, we compute FID~\cite{fid}, FVD~\cite{fvd}, and LPIPS~\cite{lpips} between the videos with inserted vehicles and the original videos for all methods.
For geometric consistency, we leverage the fact that 3D reconstruction quality depends on geometric coherence. We therefore apply the urban reconstruction method OmniRe~\cite{chen2024omnire} to the edited videos and report PSNR and SSIM~\cite{ssim} of the reconstructed results. Better geometric consistency facilitates more accurate foreground reconstruction, resulting in higher PSNR and SSIM, whereas poor consistency introduces blur and artifacts after reconstruction, leading to lower scores.
In addition, due to the longer sequences in Waymo, we further evaluate long-term generation performance on this dataset.

\begin{figure}[t]
    \centering
    \includegraphics[width=\linewidth]{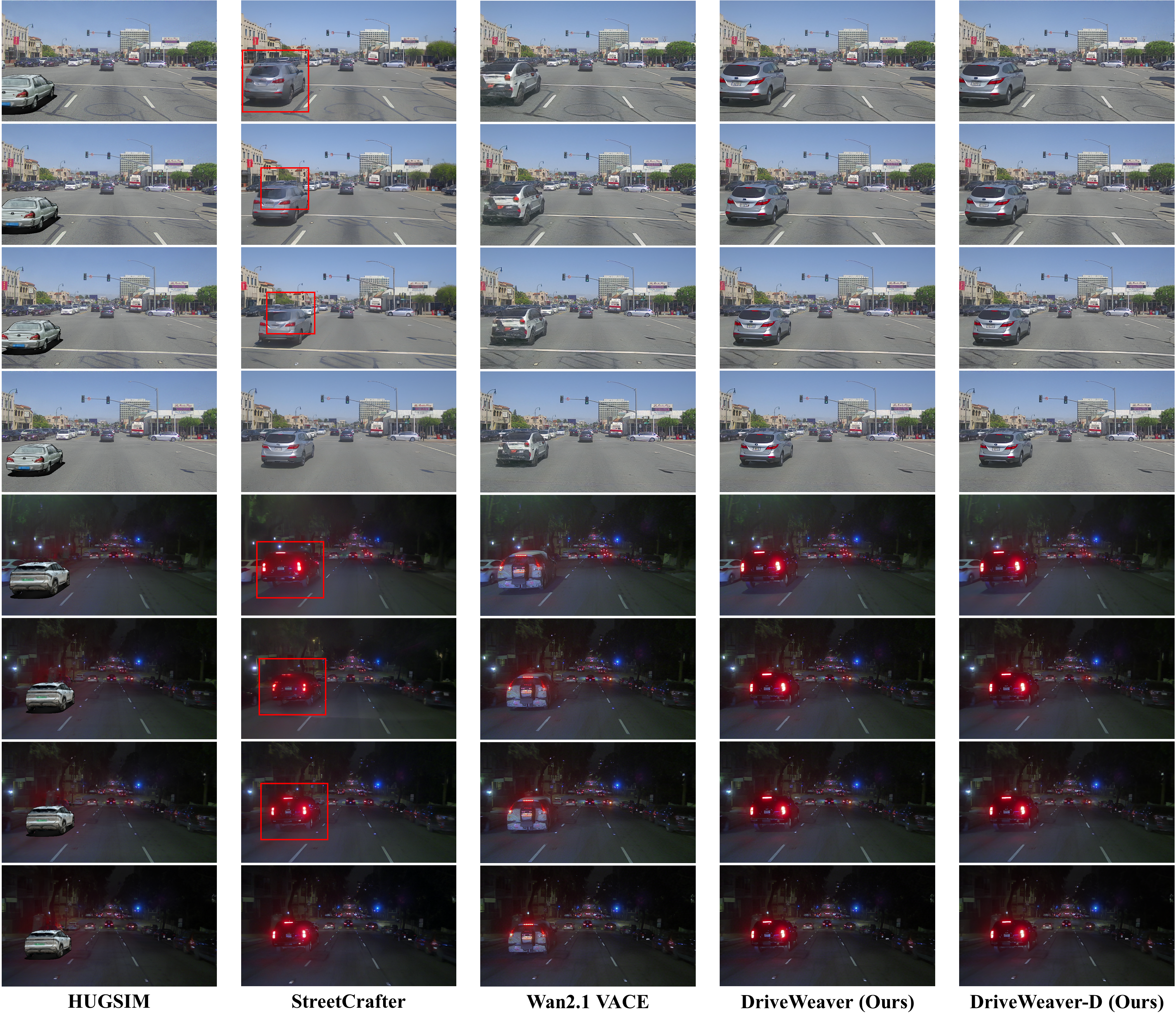}
    \caption{\textbf{Qualitative comparison on the PandaSet~\cite{pandaset} dataset.} 
    }
    \label{fig:pandaset}
\end{figure}
\subsection{Comparisons with the state-of-the-art}
\definecolor{best_result}{rgb}{0.96, 0.57, 0.58}
\definecolor{second_result}{rgb}{0.98, 0.78, 0.57}

\begin{table}[t]
    \centering
    \caption{\textbf{Comparison to state-of-the-art methods on Waymo~\cite{waymo}.} We color the top results as \colorbox{best_result}{best} and \colorbox{second_result}{second best}.}
    \resizebox{\linewidth}{!}{
    \begin{tabular}{l|cccccc|cccc}
    \toprule
    \multirow{3}{*}{\textbf{Methods}} & \multicolumn{6}{c|}{\textbf{Visual Realism}} & \multicolumn{4}{c}{\textbf{Geometric Consistency}} \\
    & \multicolumn{3}{c}{Short-term} & \multicolumn{3}{c|}{Long-term} & \multicolumn{2}{c}{Short-term} & \multicolumn{2}{c}{Long-term} \\
    & FID$\downarrow$ & FVD$\downarrow$ & LPIPS$\downarrow$ & FID$\downarrow$ & FVD$\downarrow$ & LPIPS$\downarrow$ & PSNR$\uparrow$ & SSIM$\uparrow$ & PSNR$\uparrow$ & SSIM$\uparrow$ \\
    \midrule
     Ground truth & 0 & 0 & 0 & 0 & 0 & 0 & 36.44 & 0.974 & 33.56 & 0.953 \\
    \midrule
     HUGSIM~\cite{zhou2024hugsim} & 49.06 & 325.89 & 0.304 & 56.99 & 150.95 & 0.308 & \cellcolor{best_result}36.05 & \cellcolor{best_result}0.973 & \cellcolor{best_result}33.96 & \cellcolor{best_result}0.966 \\
     StreetCrafter~\cite{yan2024streetcrafter} & 26.88 & 295.06 & 0.294 & 28.49 & 105.70 & 0.285 & 30.31 & 0.878 & 28.81 & 0.871 \\
     Wan2.1 VACE-14B~\cite{wanvace} & 54.67 & 264.74 & 0.171 & 35.12 & 63.90 & \cellcolor{second_result}0.169 & 34.47 & 0.956 & 31.99 & 0.930 \\
    \midrule
     \model{} (Ours) & 
     \cellcolor{best_result}\textbf{15.50} & 
     \cellcolor{best_result}\textbf{80.17} & 
     \cellcolor{best_result}\textbf{0.157} & 
     \cellcolor{best_result}\textbf{17.50} & 
     \cellcolor{best_result}\textbf{22.46} & 
     \cellcolor{best_result}\textbf{0.159} & 
     \cellcolor{second_result}\textbf{34.69} & 
     \cellcolor{second_result}\textbf{0.960} & 
     \cellcolor{second_result}\textbf{32.23} & 
     \cellcolor{second_result}\textbf{0.931} \\
     \model{}-D (Ours) & \cellcolor{second_result}22.35 & \cellcolor{second_result}119.36 & \cellcolor{second_result}0.168 & \cellcolor{second_result}24.28 & \cellcolor{second_result}38.50 & 0.172 & \multicolumn{4}{c}{---} \\
    \bottomrule
    \end{tabular}
    }
    \label{tab:waymo-compare}
\end{table}
\definecolor{best_result}{rgb}{0.96, 0.57, 0.58}
\definecolor{second_result}{rgb}{0.98, 0.78, 0.57}

\begin{table}[t]
    \centering
    \caption{\textbf{Comparison to state-of-the-art methods on PandaSet~\cite{pandaset}.} We color the top results as \colorbox{best_result}{best} and \colorbox{second_result}{second best}.}
    \resizebox{0.9\linewidth}{!}{
    \begin{tabular}{l|C{1.4cm}C{1.4cm}C{1.4cm}|C{2cm}C{2cm}}
    \toprule
    \multirow{2}{*}{\textbf{Methods}} & \multicolumn{3}{c|}{\textbf{Visual Realism}} & \multicolumn{2}{c}{\textbf{Geometric Consistency}} \\
    & FID$\downarrow$ & FVD$\downarrow$ & LPIPS$\downarrow$ & PSNR$\uparrow$ & SSIM$\uparrow$ \\
    \midrule
     Ground truth & 0 & 0 & 0 & 34.95 & 0.956 \\
    \midrule
     HUGSIM~\cite{zhou2024hugsim} & 56.77 & 530.58 & 0.204 & \cellcolor{best_result}34.89 & \cellcolor{best_result}0.954 \\
     StreetCrafter~\cite{yan2024streetcrafter} & 35.46 & 359.25 & 0.295 & 29.10 & 0.833  \\
     Wan2.1 VACE-14B~\cite{wanvace} & 42.74 & 309.43 & \cellcolor{second_result}0.196 & 31.11 & 0.924 \\
    \midrule
     \model{} (Ours) & 
     \cellcolor{best_result}\textbf{21.54} &
     \cellcolor{best_result}\textbf{146.15} &
     \cellcolor{best_result}\textbf{0.182} & 
     \cellcolor{second_result}\textbf{31.34} & 
     \cellcolor{second_result}\textbf{0.927} \\
     \model{}-D (Ours) & \cellcolor{second_result}25.69 & \cellcolor{second_result}208.75 & 0.203 & \multicolumn{2}{c}{---} \\
    \bottomrule
    \end{tabular}
    }
    \label{tab:pandaset-compare}
\end{table}

We first evaluate the short-term and long-term foreground insertion performance on Waymo~\cite{waymo} in terms of visual realism and geometric consistency. As presented in~\cref{tab:waymo-compare} and~\cref{fig:waymo}, \model{} outperforms state-of-the-art methods~\cite{zhou2024hugsim, yan2024streetcrafter, wanvace} in visual realism across all evaluation metrics, and ranks second in geometric consistency, surpassed only by the computer-graphics-based insertion approach HUGSIM~\cite{zhou2024hugsim}.
Meanwhile, the distilled 3DGS representation, \model{}-D, incurs only marginal degradation in visual realism compared to the original generation results.
In contrast, utilizing explicit 3D meshes, HUGSIM~\cite{zhou2024hugsim} naturally maintains structural rigidity across frames. However, without generative environmental understanding, it struggles to harmonize the asset with the target scene's complex illumination and style, which results in a severe domain gap, leading to poor visual realism metrics like FID and FVD.
StreetCrafter~\cite{yan2024streetcrafter} generates long videos in an autoregressive manner. Although it can produce visually appealing individual frames—resulting in relatively superior FID scores among the baselines—the generated vehicles exhibit noticeable semantic drift and shape deformation over long simulated trajectories. This lack of long-term temporal consistency leads to inferior FVD and LPIPS performance. More critically, temporal inconsistency breaks the geometric constraints required for 3D reconstruction, causing catastrophic failures in downstream reconstruction task, as reflected by extremely low PSNR and SSIM values.
Without explicit point cloud conditioning, Wan2.1 VACE-14B~\cite{wanvace} struggles to infer strict rigid-body dynamics from mask information alone, resulting in suboptimal performance in both visual realism and geometric consistency.

The same performance trends and the superiority of \model{} are consistently observed on the PandaSet~\cite{pandaset}, as shown in~\cref{tab:pandaset-compare} and~\cref{fig:pandaset}, demonstrating the robustness of our paradigm across datasets.
These results verify that \model{} seamlessly synthesizes realistic and environmentally coherent textures for inserted vehicles, while preserving the strict and long-term geometric consistency required for reliable autonomous driving simulation. For a more comprehensive evaluation of visual realism and geometric consistency, please refer to our supplementary material.

\subsection{Ablation study}
\begin{figure}[t]
    \centering
    \includegraphics[width=\linewidth]{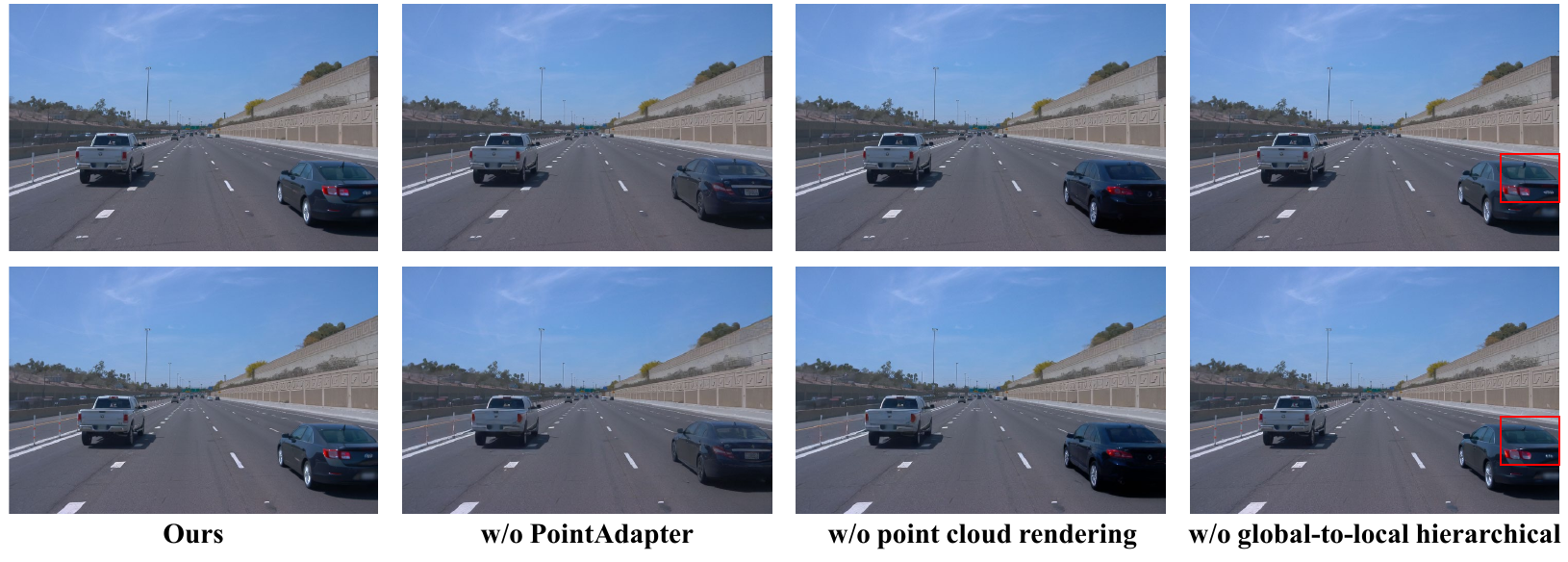}
    \caption{\textbf{Ablation study on architectural designs.} 
    }
    \label{fig:ablation}
\end{figure}

We ablate key architectural designs on the challenging Waymo long-term generation, as shown in~\cref{tab:ablation} and~\cref{fig:ablation}.

\noindent\textbf{(a) PointAdapter vs.\ full finetuning.}
Directly finetuning the Wan backbone causes FVD to more than double from 22.46 to 62.59, indicating catastrophic forgetting of the pretrained prior. The limited autonomous driving data cannot preserve the texture and motion knowledge from large-scale pretraining, confirming that lightweight adapter-based conditioning is essential for visual fidelity.

\noindent\textbf{(b) Point cloud rendering vs.\ naive projection.}
Replacing our rendering with naive 2D point projections yields overly sparse geometric conditions, causing the model to produce blurry outputs that lack high-frequency textural details. FVD degrades from 22.46 to 28.53, underscoring the importance of dense rasterized point guidance for reliable geometric supervision.

\noindent\textbf{(c) Global-to-local vs.\ autoregressive generation.}
Reverting to a standard autoregressive paradigm primarily compromises temporal coherence: FVD rises from 22.46 to 29.41 as per-step prediction error progressively accumulates over time. Without global anchor frames to enforce long-range identity and pose constraints, the inserted vehicles gradually drift in appearance and deform in shape over extended trajectories, confirming that the global-to-local hierarchical strategy is indispensable for maintaining long-term consistency.

\begin{table}[t]
\centering
\caption{\textbf{Ablation study on architectural designs.} We evaluate metrics on the challenging Waymo long-term generation.}
\resizebox{0.9\linewidth}{!}{
\begin{tabular}{clC{1.3cm}C{1.3cm}C{1.3cm}C{2cm}C{2cm}} 
\toprule
& & \multicolumn{3}{c}{\textbf{Visual Realism}} & \multicolumn{2}{c}{\textbf{Geometric Consistency}} \\
& & FID$\downarrow$ & FVD$\downarrow$ & LPIPS$\downarrow$ & PSNR$\uparrow$ & SSIM$\uparrow$  \\
\midrule
(a) & w/o PointAdapter
&  28.82 & 62.59 & 0.166 & 32.13 & 0.931 \\
(b) & w/o point cloud rendering
&  21.08 & 28.53 & 0.160 & 31.90 & 0.929 \\
(c) & w/o global-to-local hierarchical
&  18.31 & 29.41 & 0.160 & 31.89 & 0.929 \\
\midrule
& \textbf{\model{} (Ours)}
& \textbf{17.50} & \textbf{22.46} & \textbf{0.159} & \textbf{32.23} & \textbf{0.931} \\
\bottomrule
\end{tabular}
}
\label{tab:ablation}
\end{table}
\section{Conclusion}
We present \model{}, a novel point-conditioned video inpainting diffusion model designed for controllable vehicle insertion in autonomous driving simulation. By conditioning the inpainting process on rendered vehicle point cloud maps, the model generates high-quality foreground vehicles that are both environmentally harmonious and geometrically consistent. To support extended durations, we introduce a global-to-local hierarchical inpainting strategy, which ensures the long-term temporal consistency of the inserted objects. Furthermore, by distilling the generated foreground into 3D Gaussian representations, our framework facilitates the real-time rendering requirements of autonomous driving simulators. Extensive comparative evaluations across multiple datasets demonstrate the effectiveness of the proposed method, providing a robust tool for scalable scene augmentation in autonomous driving simulation.


\section*{Acknowledgements}
This work was supported in part by New Generation Artificial Intelligence-National Science and Technology Major Project (2025ZD0123004), Ningbo grant (2025Z038) and National Natural Science Foundation of China (Grant No.623760 60)

%
%
\bibliographystyle{splncs04}
\bibliography{main}

\begin{thebibliography}{10}
\providecommand{\url}[1]{\texttt{#1}}
\providecommand{\urlprefix}{URL }
\providecommand{\doi}[1]{https://doi.org/#1}

\bibitem{chen2023seine}
Chen, X., Wang, Y., Zhang, L., Zhuang, S., Ma, X., Yu, J., Wang, Y., Lin, D., Qiao, Y., Liu, Z.: Seine: Short-to-long video diffusion model for generative transition and prediction. In: ICLR (2023)

\bibitem{chen2023periodic}
Chen, Y., Gu, C., Jiang, J., Zhu, X., Zhang, L.: Periodic vibration gaussian: Dynamic urban scene reconstruction and real-time rendering. arXiv preprint  (2023)

\bibitem{chen2025snerf}
Chen, Y., Zhang, J., Xie, Z., Li, W., Zhang, F., Lu, J., Zhang, L.: S-nerf++: Autonomous driving simulation via neural reconstruction and generation. IEEE TPAMI  (2025)

\bibitem{chen2024omnire}
Chen, Z., Yang, J., Huang, J., de~Lutio, R., Esturo, J.M., Ivanovic, B., Litany, O., Gojcic, Z., Fidler, S., Pavone, M., Song, L., Wang, Y.: Omnire: Omni urban scene reconstruction. arXiv preprint  (2024)

\bibitem{chung2023unimax}
Chung, H.W., Constant, N., Garcia, X., Roberts, A., Tay, Y., Narang, S., Firat, O.: Unimax: Fairer and more effective language sampling for large-scale multilingual pretraining. arXiv preprint  (2023)

\bibitem{dosovitskiy2017carla}
Dosovitskiy, A., Ros, G., Codevilla, F., Lopez, A., Koltun, V.: Carla: An open urban driving simulator. In: CoRL (2017)

\bibitem{du20243drealcar}
Du, X., Wang, Y., Sun, H., Wu, Z., Sheng, H., Wang, S., Ying, J., Lu, M., Zhu, T., Zhan, K., Yu, X.: 3drealcar: An in-the-wild rgb-d car dataset with 360-degree views. In: ICCV (2025)

\bibitem{esser2024sd3}
Esser, P., Kulal, S., Blattmann, A., Entezari, R., M{\"u}ller, J., Saini, H., Levi, Y., Lorenz, D., Sauer, A., Boesel, F., et~al.: Scaling rectified flow transformers for high-resolution image synthesis. In: ICML (2024)

\bibitem{gao2025lora}
Gao, C., Ding, L., Cai, X., Huang, Z., Wang, Z., Xue, T.: Lora-edit: Controllable first-frame-guided video editing via mask-aware lora fine-tuning. arXiv preprint  (2025)

\bibitem{RAD}
Gao, H., Chen, S., Jiang, B., Liao, B., Shi, Y., Guo, X., Pu, Y., Yin, H., Li, X., Zhang, X., Zhang, Y., Liu, W., Zhang, Q., Wang, X.: Rad: Training an end-to-end driving policy via large-scale 3dgs-based reinforcement learning. arXiv preprint  (2025)

\bibitem{gao2024vista}
Gao, S., Yang, J., Chen, L., Chitta, K., Qiu, Y., Geiger, A., Zhang, J., Li, H.: Vista: A generalizable driving world model with high fidelity and versatile controllability. In: NIPS (2024)

\bibitem{guo2023animatediff}
Guo, Y., Yang, C., Rao, A., Wang, Y., Qiao, Y., Lin, D., Dai, B.: Animatediff: Animate your personalized text-to-image diffusion models without specific tuning. arXiv preprint  (2023)

\bibitem{nuplan}
H.~Caesar, J.~Kabzan, K.T.e.a.: Nuplan: A closed-loop ml-based planning benchmark for autonomous vehicles. In: CVPR ADP3 workshop (2021)

\bibitem{henschel2024streamingt2v}
Henschel, R., Khachatryan, L., Hayrapetyan, D., Poghosyan, H., Tadevosyan, V., Wang, Z., Navasardyan, S., Shi, H.: Streamingt2v: Consistent, dynamic, and extendable long video generation from text. In: CVPR (2025)

\bibitem{fid}
Heusel, M., Ramsauer, H., Unterthiner, T., Nessler, B., Hochreiter, S.: Gans trained by a two time-scale update rule converge to a local nash equilibrium. In: NIPS (2017)

\bibitem{ddpm}
Ho, J., Jain, A., Abbeel, P.: Denoising diffusion probabilistic models. arXiv preprint  (2020)

\bibitem{cfg}
Ho, J., Salimans, T.: Classifier-free diffusion guidance. arXiv preprint  (2022)

\bibitem{hu2020proposal}
Hu, Y.T., Wang, H., Ballas, N., Grauman, K., Schwing, A.G.: Proposal-based video completion. In: ECCV (2020)

\bibitem{realengine}
Jiang, J., Song, N., Li, J., Zhu, X., Zhang, L.: Realengine: Simulating autonomous driving in realistic context. arXiv preprint  (2025)

\bibitem{wanvace}
Jiang, Z., Han, Z., Mao, C., Zhang, J., Pan, Y., Liu, Y.: Vace: All-in-one video creation and editing. arXiv preprint  (2025)

\bibitem{kerbl3Dgaussians}
Kerbl, B., Kopanas, G., Leimk{\"u}hler, T., Drettakis, G.: 3d gaussian splatting for real-time radiance field rendering. In: ACM TOG (2023)

\bibitem{kong2024hunyuanvideo}
Kong, W., Tian, Q., Zhang, Z., Min, R., Dai, Z., Zhou, J., Xiong, J., Li, X., Wu, B., Zhang, J., et~al.: Hunyuanvideo: A systematic framework for large video generative models. arXiv preprint  (2024)

\bibitem{g2editor}
Li, J., Jiang, J., Miao, J., Long, M., Wen, T., Jia, P., Liu, S., Yu, C., Liu, M., Cai, Y., et~al.: Realistic and controllable 3d gaussian-guided object editing for driving video generation. arXiv preprint  (2025)

\bibitem{dipir}
Liang, R., Gojcic, Z., Nimier-David, M., Acuna, D., Vijaykumar, N., Fidler, S., Wang, Z.: Photorealistic object insertion with diffusion-guided inverse rendering. In: ECCV (2024)

\bibitem{driveeditor}
Liang, Y., Yan, Z., Chen, L., Zhou, J., Yan, L., Zhong, S., Zou, X.: Driveeditor: A unified 3d information-guided framework for controllable object editing in driving scenes. In: AAAI (2025)

\bibitem{flow-matching}
Lipman, Y., Chen, R.T.Q., Ben-Hamu, H., Nickel, M., Le, M.: Flow matching for generative modeling. In: ICLR (2023)

\bibitem{r3d2}
Ljungbergh, W., Taveira, B., Zheng, W., Tonderski, A., Peng, C., Kahl, F., Petersson, C., Felsberg, M., Keutzer, K., Tomizuka, M., et~al.: R3d2: Realistic 3d asset insertion via diffusion for autonomous driving simulation. In: CVPR Workshops (2026)

\bibitem{ljungbergh2024neuroncap}
Ljungbergh, W., Tonderski, A., Johnander, J., Caesar, H., {\AA}str{\"o}m, K., Felsberg, M., Petersson, C.: Neuroncap: Photorealistic closed-loop safety testing for autonomous driving. In: ECCV (2024)

\bibitem{lu2024urbancad}
Lu, Y., Cai, Y., Zhang, S., Zhou, H., Hu, H., Yu, H., Geiger, A., Liao, Y.: Urbancad: Towards highly controllable and photorealistic 3d vehicles for urban scene simulation. In: CVPR (2025)

\bibitem{Ma2025BezierGS}
Ma, Z., Jiang, J., Chen, Y., Zhang, L.: Béziergs: Dynamic urban scene reconstruction with bézier curve gaussian splatting. In: ICCV (2025)

\bibitem{mildenhall2020nerf}
Mildenhall, B., Srinivasan, P.P., Tancik, M., Barron, J.T., Ramamoorthi, R., Ng, R.: Nerf: Representing scenes as neural radiance fields for view synthesis. In: ECCV (2020)

\bibitem{nsg}
Ost, J., Mannan, F., Thuerey, N., Knodt, J., Heide, F.: Neural scene graphs for dynamic scenes. In: CVPR (2021)

\bibitem{Peebles2022DiT}
Peebles, W., Xie, S.: Scalable diffusion models with transformers. In: ICCV (2023)

\bibitem{qiu2025cinescale}
Qiu, H., Yu, N., Huang, Z., Debevec, P., Liu, Z.: Cinescale: Free lunch in high-resolution cinematic visual generation. arXiv preprint  (2025)

\bibitem{sd}
Rombach, R., Blattmann, A., Lorenz, D., Esser, P., Ommer, B.: High-resolution image synthesis with latent diffusion models. In: CVPR (2022)

\bibitem{airsim2017fsr}
Shah, S., Dey, D., Lovett, C., Kapoor, A.: Airsim: High-fidelity visual and physical simulation for autonomous vehicles. In: FSR (2017)

\bibitem{genmm}
Singh, B., Kulharia, V., Yang, L., Ravichandran, A., Tyagi, A., Shrivastava, A.: Genmm: Geometrically and temporally consistent multimodal data generation for video and lidar. arXiv preprint  (2024)

\bibitem{waymo}
Sun, P., Kretzschmar, H., Dotiwalla, X., Chouard, A., Patnaik, V., Tsui, P., Guo, J., Zhou, Y., Chai, Y., Caine, B., et~al.: Scalability in perception for autonomous driving: Waymo open dataset. In: CVPR (2020)

\bibitem{turki2023suds}
Turki, H., Zhang, J.Y., Ferroni, F., Ramanan, D.: Suds: Scalable urban dynamic scenes. arXiv preprint  (2023)

\bibitem{fvd}
Unterthiner, T., van Steenkiste, S., Kurach, K., Marinier, R., Michalski, M., Gelly, S.: Towards accurate generative models of video: A new metric \& challenges. arXiv preprint  (2018)

\bibitem{wan}
Wan, T., Wang, A., Ai, B., Wen, B., Mao, C., Xie, C.W., Chen, D., Yu, F., Zhao, H., Yang, J., Zeng, J., Wang, J., Zhang, J., Zhou, J., Wang, J., Chen, J., Zhu, K., Zhao, K., Yan, K., Huang, L., Feng, M., Zhang, N., Li, P., Wu, P., Chu, R., Feng, R., Zhang, S., Sun, S., Fang, T., Wang, T., Gui, T., Weng, T., Shen, T., Lin, W., Wang, W., Wang, W., Zhou, W., Wang, W., Shen, W., Yu, W., Shi, X., Huang, X., Xu, X., Kou, Y., Lv, Y., Li, Y., Liu, Y., Wang, Y., Zhang, Y., Huang, Y., Li, Y., Wu, Y., Liu, Y., Pan, Y., Zheng, Y., Hong, Y., Shi, Y., Feng, Y., Jiang, Z., Han, Z., Wu, Z.F., Liu, Z.: Wan: Open and advanced large-scale video generative models. arXiv preprint  (2025)

\bibitem{wang2019video}
Wang, C., Huang, H., Han, X., Wang, J.: Video inpainting by jointly learning temporal structure and spatial details. In: AAAI (2019)

\bibitem{wang2024freevs}
Wang, Q., Fan, L., Wang, Y., Chen, Y., Zhang, Z.: Freevs: Generative view synthesis on free driving trajectory. In: ICLR (2025)

\bibitem{ssim}
Wang, Z., Bovik, A.C., Sheikh, H.R., Simoncelli, E.P.: Image quality assessment: from error visibility to structural similarity. TIP  (2004)

\bibitem{wang2023fegr}
Wang, Z., Shen, T., Gao, J., Huang, S., Munkberg, J., Hasselgren, J., Gojcic, Z., Chen, W., Fidler, S.: Neural fields meet explicit geometric representations for inverse rendering of urban scenes. In: CVPR (2023)

\bibitem{pandaset}
Xiao, P., Shao, Z., Hao, S., Zhang, Z., Chai, X., Jiao, J., Li, Z., Wu, J., Sun, K., Jiang, K., Wang, Y., Yang, D.: Pandaset: Advanced sensor suite dataset for autonomous driving. In: ITSC (2021)

\bibitem{xie2023snerf}
Xie, Z., Zhang, J., Li, W., Zhang, F., Zhang, L.: S-nerf: Neural radiance fields for street views. In: ICLR (2023)

\bibitem{Xu_2019_CVPR}
Xu, R., Li, X., Zhou, B., Loy, C.C.: Deep flow-guided video inpainting. In: CVPR (2019)

\bibitem{yan2024street}
Yan, Y., Lin, H., Zhou, C., Wang, W., Sun, H., Zhan, K., Lang, X., Zhou, X., Peng, S.: Street gaussians: Modeling dynamic urban scenes with gaussian splatting. In: ECCV (2024)

\bibitem{yan2024streetcrafter}
Yan, Y., Xu, Z., Lin, H., Jin, H., Guo, H., Wang, Y., Zhan, K., Lang, X., Bao, H., Zhou, X., Peng, S.: Streetcrafter: Street view synthesis with controllable video diffusion models. In: CVPR (2025)

\bibitem{yang2024storm}
Yang, J., Huang, J., Chen, Y., Wang, Y., Li, B., You, Y., Sharma, A., Igl, M., Karkus, P., Xu, D., et~al.: Storm: Spatio-temporal reconstruction model for large-scale outdoor scenes. arXiv preprint  (2024)

\bibitem{yang2023emernerf}
Yang, J., Ivanovic, B., Litany, O., Weng, X., Kim, S.W., Li, B., Che, T., Xu, D., Fidler, S., Pavone, M., Wang, Y.: Emernerf: Emergent spatial-temporal scene decomposition via self-supervision. arXiv preprint  (2023)

\bibitem{yang2023unisim}
Yang, Z., Chen, Y., Wang, J., Manivasagam, S., Ma, W.C., Yang, A.J., Urtasun, R.: Unisim: A neural closed-loop sensor simulator. In: CVPR (2023)

\bibitem{yang2024drivex}
Yang, Z., Pan, Z., Yang, Y., Zhu, X., Zhang, L.: Driving scene synthesis on free-form trajectories with generative prior. In: ICCV (2025)

\bibitem{yang2024cogvideox}
Yang, Z., Teng, J., Zheng, W., Ding, M., Huang, S., Xu, J., Yang, Y., Hong, W., Zhang, X., Feng, G., et~al.: Cogvideox: Text-to-video diffusion models with an expert transformer. arXiv preprint  (2024)

\bibitem{lpips}
Zhang, R., Isola, P., Efros, A.A., Shechtman, E., Wang, O.: The unreasonable effectiveness of deep features as a perceptual metric. In: CVPR (2018)

\bibitem{lionlora}
Zhang, Y., Cao, C., Yu, C., Zhu, J.: Lion-lora: Rethinking lora fusion to unify controllable spatial and temporal generation for video diffusion. In: ICCV (2025)

\bibitem{zhao2024drivedreamer4d}
Zhao, G., Ni, C., Wang, X., Zhu, Z., Zhang, X., Wang, Y., Huang, G., Chen, X., Wang, B., Zhang, Y., Mei, W., Wang, X.: Drivedreamer4d: World models are effective data machines for 4d driving scene representation. In: CVPR (2025)

\bibitem{zheng2026versecrafter}
Zheng, S., Yin, M., Hu, W., Li, X., Shan, Y., Fu, Y.: Versecrafter: Dynamic realistic video world model with 4d geometric control. arXiv preprint  (2026)

\bibitem{opensora}
Zheng, Z., Peng, X., Yang, T., Shen, C., Li, S., Liu, H., Zhou, Y., Li, T., You, Y.: Open-sora: Democratizing efficient video production for all. arXiv preprint  (2024)

\bibitem{zhou2024hugsim}
Zhou, H., Lin, L., Wang, J., Lu, Y., Bai, D., Liu, B., Wang, Y., Geiger, A., Liao, Y.: Hugsim: A real-time, photo-realistic and closed-loop simulator for autonomous driving. arXiv preprint  (2024)

\bibitem{hugs}
Zhou, H., Shao, J., Xu, L., Bai, D., Qiu, W., Liu, B., Wang, Y., Geiger, A., Liao, Y.: Hugs: Holistic urban 3d scene understanding via gaussian splatting. In: CVPR (2024)

\bibitem{propainter}
Zhou, S., Li, C., Chan, K.C., Loy, C.C.: {ProPainter}: Improving propagation and transformer for video inpainting. In: ICCV (2023)

\bibitem{zhou2025nexus}
Zhou, Y., Ye, N., Ljungbergh, W., Li, T., Yang, J., Yang, Z., Zhu, H., Petersson, C., Li, H.: Decoupled diffusion sparks adaptive scene generation. In: ICCV (2025)

\bibitem{zou2021progressive}
Zou, X., Yang, L., Liu, D., Lee, Y.J.: Progressive temporal feature alignment network for video inpainting. In: CVPR (2021)

\end{thebibliography}
\end{document}